\documentclass[11pt]{article}

\usepackage[preprint]{acl}

\usepackage{fontawesome5}

\usepackage{times}
\usepackage{latexsym}
\usepackage{booktabs}
\usepackage{amsmath}
\usepackage{amssymb}
\usepackage{float}
\usepackage{footmisc}

\usepackage{hyperref}
\usepackage{xurl} 

\usepackage[T1]{fontenc}

\usepackage[utf8]{inputenc}
\usepackage[table]{xcolor}
\usepackage{booktabs}

\usepackage{colortbl}
\usepackage{xcolor}
\usepackage{array}
\usepackage{pgf}

\definecolor{headerblue}{RGB}{30, 77, 140}
\definecolor{rowgray}{RGB}{245, 245, 245}
\definecolor{bestgreen}{RGB}{220, 240, 220}
\usepackage{microtype}

\usepackage{inconsolata}

\usepackage{graphicx}
\usepackage{amsmath} 

 \usepackage{multirow}

\NewDocumentCommand{\bx}
{ mO{} }{\textcolor{blue}{\textsuperscript{\textit{bx}}\textsf{\textbf{\small[#1]}}}}

\NewDocumentCommand{\heng}
{ mO{} }{\textcolor{green}{\textsuperscript{\textit{heng}}\textsf{\textbf{\small[#1]}}}}

\NewDocumentCommand{\xy}
{ mO{} }{\textcolor{red}{\textsuperscript{\textit{xy}}\textsf{\textbf{\small[#1]}}}}

\NewDocumentCommand{\js}
{ mO{} }{\textcolor{orange}{\textsuperscript{\textit{js}}\textsf{\textbf{\small[#1]}}}}

\newcommand{\Induce}{\mathrm{Induce}}
\newcommand{\Execute}{\mathrm{Execute}}
\newcommand{\Refine}{\mathrm{Refine}}

\newcommand{\ModelName}{\textsc{PaperPilot}}
\newcommand{\Toolset}{\ModelName{}\textit{\textsc{-Toolset}}}
\newcommand{\Workflow}{\ModelName{}\textit{\textsc{-Workflow}}}

\newcommand{\minlogce}{-0.928}  
\newcommand{\maxlogce}{ 2.226}  

\newcommand{\cecolor}[1]{%
  \pgfmathtruncatemacro{\norm}{max(0,min(100,#1))}%
  \pgfmathtruncatemacro{\inv}{100-\norm}%
  \edef\clr{red!\inv!green!70!black}%
  \expandafter\color\expandafter{\clr}%
}

\newcommand{\ceraw}[1]{%
  \pgfmathsetmacro{\logv}{ln(#1)/ln(10)}%
  \pgfmathsetmacro{\normraw}{100*(\logv-\minlogce)/(\maxlogce-\minlogce)}%
  {\bfseries\cecolor{\normraw}#1}%
}

\newcommand{\releff}[1]{%
  \pgfmathsetmacro{\normraw}{100*(#1-0.47)/(3.03-0.47)}%
  \pgfmathtruncatemacro{\norm}{max(0,min(100,\normraw))}%
  \pgfmathtruncatemacro{\inv}{100-\norm}%
  \edef\clr{red!\inv!green!70!black}%
  {\bfseries\expandafter\color\expandafter{\clr}#1}%
}

\definecolor{paperpilotpink}{HTML}{CC79A7}

\title{Multi-Turn Agentic Scientific Literature Search via Workflow Induction}

\author{
Jisen Li\textsuperscript{1,2}\thanks{Equal contribution, order interchangeable.}\thanks{Contact: \texttt{\{jisenli2, bl61\}@illinois.edu}} \quad
Bingxuan Li\textsuperscript{1}\footnotemark[1] \quad
Nanyi Jiang\textsuperscript{3}\footnotemark[1] \\
\textbf{Xuying Ning}\textsuperscript{1} \quad
\textbf{Xiyao Wang}\textsuperscript{3} \quad
\textbf{Yifan Shen}\textsuperscript{1} \quad
\textbf{Heng Wang}\textsuperscript{1} \quad
\textbf{Yuqing Jian}\textsuperscript{2}
\\
\textbf{Xiaoxia Wu}\textsuperscript{2} \quad
\textbf{Ben Athiwaratkun}\textsuperscript{2} \quad
\textbf{Pan Lu}\textsuperscript{4} \quad
\textbf{Jiaxuan You}\textsuperscript{1} \quad
\textbf{Bingxin Zhao}\textsuperscript{3}
\\[0.6em]
\textsuperscript{1}University of Illinois Urbana-Champaign \quad
\textsuperscript{2}Together AI \\
\textsuperscript{3}University of Pennsylvania \quad
\textsuperscript{4}Stanford University 
\\[0.4em]
\begin{tabular}{c}
\small
\faGlobe\quad \href{https://paperpilot.papersearch.org/}{Project Website}
\quad
\faGithub\quad \href{https://github.com/mtilyxuegao/PaperPilot}{Code}
\end{tabular}
}

\begin{document}
\maketitle

\begin{abstract}
Scientific literature search often requires more than retrieving papers from a single query: users' intents are underspecified, preference-dependent, and evolve through interaction. Existing search agents typically rely on fixed pipelines or implicit language-only reasoning, making their search strategies difficult to control, inspect, and refine. We introduce \ModelName{}, a multi-turn literature search agent that frames scientific search as workflow induction. Given an anchor paper and a user query, \ModelName{} constructs an executable DAG of paper-search operators, including keyword search, citation expansion, filtering, scoring, reranking, and evidence extraction. User feedback is then used to refine both the query and the workflow itself. We train \ModelName{} with supervised workflow imitation and preference optimization over controlled workflow corruptions. Experiments show that \ModelName{}-9B improves over the base Qwen3.5-9B toolset agent under multi-turn interaction, increasing Hit@5 from 58.0 to 77.0, MRR from 47.5 to 59.4, and nDCG@10 from 26.8 to 32.5, while reducing workflow execution errors from 9.5\% to 0\%. These results show that explicit, editable search workflows provide an effective and controllable interface for aligning literature search agents with complex scientific intent.   
\end{abstract}

\begin{figure}
    \centering
    \includegraphics[width=0.89\linewidth]{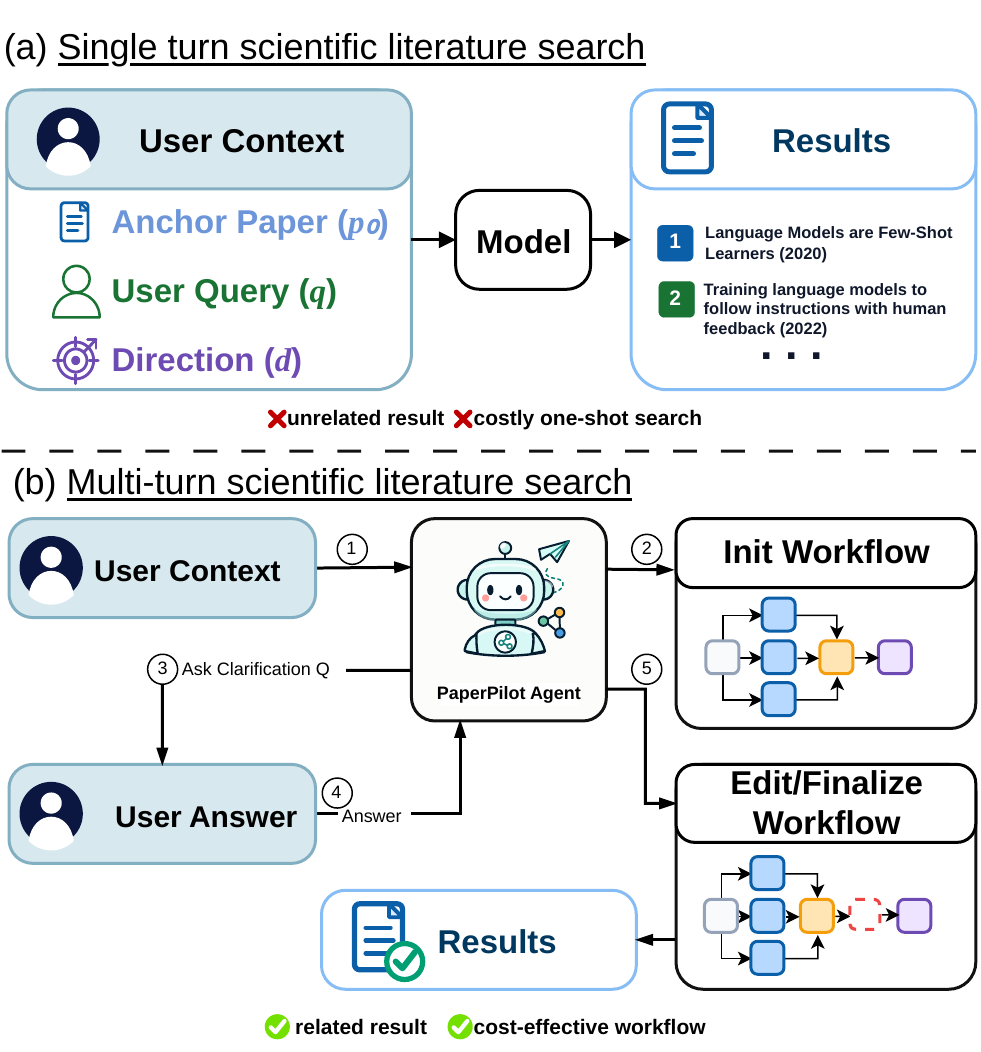}
    \caption{
Overview of \ModelName{}.
Compared with single-turn scientific literature search, \ModelName{} uses multi-turn feedback to clarify user intent and refine the search workflow before producing final results.
}
    \label{fig:placeholder}
    \vspace{-0.2in}
\end{figure}

\begin{table*}[t]
\centering
\small
\setlength{\tabcolsep}{6pt}
\renewcommand{\arraystretch}{0.8}
\begin{tabular}{lccccc}
\toprule
\textbf{System}
& \shortstack{\textbf{Symbolic}\\\textbf{Workflow}}
& \shortstack{\textbf{Workflow}\\\textbf{Refinement}}
& \shortstack{\textbf{Multi-turn}\\\textbf{Dialogue}}
& \shortstack{\textbf{Citation}\\\textbf{Expansion}}
& \shortstack{\textbf{Evidence}\\\textbf{Grounding}} \\
\midrule
OpenAI DeepResearch \citep{openai2025deepresearch} & $\times$ & $\times$ & $\triangle$ & $-$         & $\triangle$ \\
LitLLM \citep{agarwal2024litllm}            & $\times$    & $\times$    & $\times$    & $\times$    & $\times$    \\
STORM \citep{shao2024assisting}                 & $\times$    & $\times$    & $\times$    & $\times$    & $\triangle$ \\
PaperQA2 \citep{skarlinski2024language}           & $\times$    & $\times$    & $\times$    & $\checkmark$& $\checkmark$\\
OpenScholar \citep{asai2024openscholar}     & $\times$    & $\times$    & $\times$    & $\times$    & $\triangle$ \\
AutoSurvey \citep{wang2024autosurvey}       & $\times$    & $\times$    & $\times$    & $\times$    & $\triangle$ \\
ResearchAgent \citep{baek2025researchagent} & $\times$    & $\times$    & $\times$    & $\triangle$ & $\times$    \\
ChatCite \citep{li2024chatcite}             & $\times$    & $\times$    & $\times$    & $\times$    & $\triangle$ \\
AI Scientist \citep{lu2024aiscientist}      & $\times$    & $\times$    & $\times$    & $\triangle$ & $\times$    \\
PaSa \citep{he2025pasa}                     & $\times$    & $\times$    & $\times$    & $\checkmark$& $\times$    \\
SPAR \citep{li2025spar}                     & $\times$    & $\times$    & $\times$    & $\triangle$ & $\times$    \\
Elicit \citep{whitfield2023elicit}          & $\times$    & $\times$    & $\times$    & $\times$    & $\triangle$ \\
\textbf{\ModelName{} (Ours)}                & $\checkmark$& $\checkmark$& $\checkmark$& $\checkmark$& $\checkmark$\\
\bottomrule
\end{tabular}
\caption{Comparison of literature search and agentic reasoning systems by key capabilities.
$\checkmark$ = explicit first-class support; $\triangle$ = partial, or indirect; $\times$ = clearly unsupported.
}
\vspace{-0.2in}
\label{tab:related_work_comparison}
\end{table*}

\section{Introduction}

Scientific literature search is a core research activity that increasingly requires more than keyword matching. 
Researchers must identify relevant prior work, explore citation and semantic neighborhoods, compare candidate papers, filter distractors, and understand why retrieved papers are relevant to a given research intent. 
As scientific output continues to grow, literature search increasingly depends on adaptive retrieval and interaction rather than static query matching.

Recent agentic search systems improve retrieval quality by combining language models with external tools, iterative reasoning, and multi-step search~\citep{agarwal2024litllm,skarlinski2024language,asai2024openscholar,he2025pasa,baek2025researchagent}. 
However, search intent is rarely fully specified in a single query, echoing broader findings in clarification-driven retrieval that ambiguous information needs often require interactive disambiguation~\citep{zamani2020generating,metal, chi2024clarinet,liu2026adaplanbench, li2026pearl, wang2026bioinsight}.  
For example, a request such as ``find follow-up work on this paper'' may refer to direct citations, recent extensions, papers in the same application domain, or work building upon a specific methodological component. 
The correct retrieval strategy therefore depends on latent user preferences and evolving interaction feedback.

This motivates a multi-turn setting for scientific literature search, where the agent interacts with the user to clarify intent and refine the retrieval strategy before finalizing results. 
Such interaction is important because scientific relevance often depends on fine-grained preferences, such as citation direction, methodological similarity, recency, benchmark usage, or application domain.

Most existing AI agents represent human user intent implicitly in free-form language or execute largely fixed retrieval-and-synthesis pipelines~\citep{agarwal2024litllm,shao2024assisting,li-etal-2024-control,wang2024autosurvey,asai2024openscholar,he2025pasa,echofoley,liu2026naacl}, making the retrieval process difficult to control. 
When users provide feedback such as ``these papers are too broad'' or ``focus on more recent work,'' the agent must translate natural-language feedback into concrete retrieval changes. 
Without an effective representation, such feedback is often treated as additional query text rather than as an instruction to modify retrieval behavior itself.

We propose \ModelName{}, a multi-turn agentic scientific literature search system via workflow induction. Instead of relying on a fixed retrieval pipeline, \ModelName{} constructs an executable directed acyclic graph (DAG) for each user request. We define \ModelName{}-Toolset as a library of typed paper-search operators, including keyword search, citation expansion, filtering, scoring, reranking, and evidence extraction. 
Each edge represents the flow of intermediate outputs such as paper sets, keywords, evidence, or ranking scores.
This representation allows the agent to compose intent-specific search strategies and directly refine them through user feedback. 
For example, a query about emerging follow-up work can emphasize recent keyword search and forward citation expansion, while a query about strong baselines can prioritize citation-based expansion and benchmark filtering. To optimize the performance, we train \ModelName{}-9B  model using a two-stage supervision pipeline.  We first collect high-quality workflow trajectories from a strong teacher model across five search directions: predecessor, successor, sibling, benchmark, and survey.  We then construct preference pairs by corrupting successful workflows with common structural and semantic errors, and optimize the model with supervised fine-tuning followed by preference optimization.

We evaluate \ModelName{}-9B on a hold-out evaluation set, where each case fixes the anchor paper, user query, search direction, hidden gold-paper set, and interaction protocol to enable reproducible comparison across systems. 
We evaluate both single-turn and multi-turn settings: in the single-turn setting, the agent searches directly from the original query; in the multi-turn setting, an LLM-based user simulator provides clarification feedback under a leakage-control protocol.

Experiments show that \ModelName{}-9B improves over the base Qwen3.5-9B toolset agent under multi-turn interaction, increasing Hit@5 from 58.0 to 77.0, MRR from 47.5 to 59.4, and nDCG@10 from 26.8 to 32.5, while reducing workflow execution errors from 9.5\% to 0\%. These results show that explicit, editable search workflows provide an effective interface for aligning literature search agents with complex scientific intent.

In summary, our contributions are twofold. 
First, we formulate multi-turn scientific literature search as workflow induction, where user feedback is converted into edits over executable DAG-structured search workflows. 
Second, we introduce and train \ModelName{}-9B with workflow imitation and preference optimization, showing improved retrieval quality, workflow stability, and cost-efficiency on a reproducible multi-turn paper-search benchmark.
\section{\ModelName{}}

\begin{figure*}[t]
    \centering
    \includegraphics[width=\linewidth]{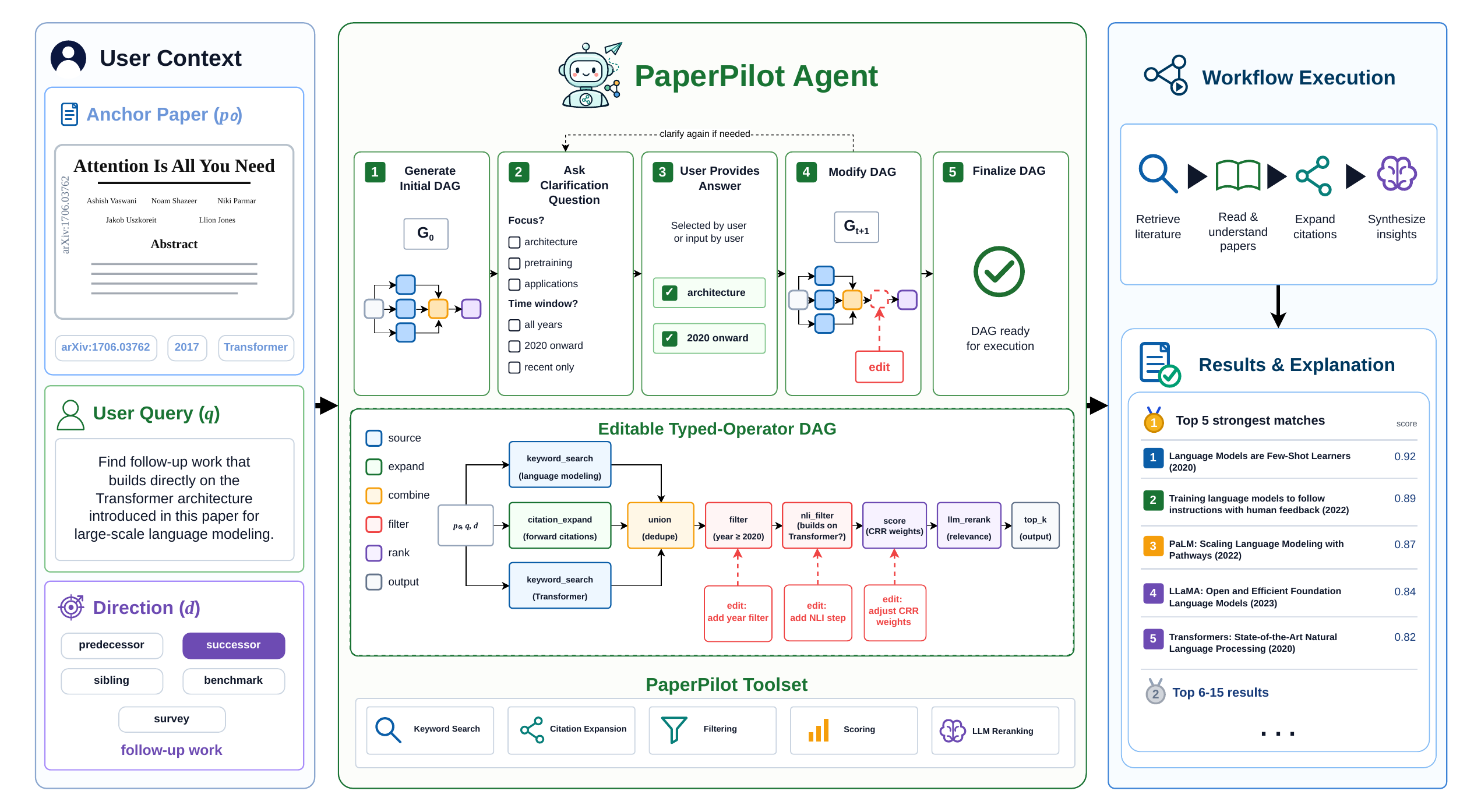}
     \vspace{-0.2in}
    \caption{Overview of \ModelName{}. Given an anchor paper and a user search intent, the agent induces a DAG-structured search workflow from a predefined toolset, executes the workflow over the literature corpus, and refines the workflow through multi-turn user interaction.
    }
    \label{fig:method}
    \vspace{-0.2in}
\end{figure*}

\subsection{Task Formulation}
We formulate multi-turn agentic paper search as an interactive workflow-construction problem. 
Given an anchor paper $p_0$ and a user query $q$, the agent aims to retrieve a ranked list of papers that satisfies the user's evolving search intent, such as finding follow-up work, identifying strong baselines, comparing related methods, or exploring adjacent research directions.

At each turn, the agent observes the current search context, including the anchor paper, the user query, the candidate paper set, and the interaction history.  It then either asks a clarification question or applies paper-search operators to expand, filter, score, rerank, or explain candidate papers. After a finite number of interaction turns, the agent outputs a ranked paper list $\hat{\mathcal{P}}$, optionally accompanied by evidence snippets or a relation graph. 
The objective is to maximize paper relevance, coverage of the requested search intent, alignment with user feedback, and interaction efficiency. 
A full formalization of the state, action space, transition function, and utility objective is provided in Appendix~\ref{app:task_formulation}.

\subsection{Paper Search as Workflow Induction}

Instead of relying on a fixed retrieval pipeline, \ModelName{} treats paper search as a workflow induction problem. Given the query $q$ and anchor paper $p_0$, the agent constructs an executable search workflow by selecting tools from a predefined toolset, configuring their parameters, and composing them into a DAG. Each node in the DAG corresponds to an operator invocation, while each edge represents the flow of intermediate outputs, such as paper sets, scores, keywords, evidence, or relation graphs.

This formulation gives the agent two forms of flexibility. First, it can choose which operators are relevant for the current search intent. For example, a query asking for ``strong baselines'' may emphasize citation expansion and fine-grained reranking, while a query asking for ``emerging follow-up work'' may rely more heavily on recent keyword search and citation tracing. Second, the agent can adapt operator parameters, such as search keywords, citation direction, filtering predicates, scoring formulas, and reranking criteria, rather than executing a one-size-fits-all pipeline.

\noindent\textbf{Toolset.}
We define a library of paper-search operators, denoted as $\Toolset{}$. Each operator has a typed input and output signature, enabling the agent to compose operators into valid workflows. The toolset covers sourcing, set operations, filtering, scoring, reranking, keyword generation, evidence extraction, and workflow construction. The detail of $\Toolset{}$ is attached to Appendix ~\ref{app:toolset}.

\noindent\textbf{Workflow representation.}
A workflow is represented as a DAG:
$G = (\mathcal{V}, \mathcal{E})$,
where each node $v_i \in \mathcal{V}$ is an instantiated operator
\[
v_i = (o_i, \theta_i), \quad o_i \in \Toolset{},
\]
and $\theta_i$ denotes its parameters. An edge $(v_i, v_j) \in \mathcal{E}$ indicates that the output of node $v_i$ is used as an input to node $v_j$. We require workflows to be type-consistent: the output type of each predecessor must match the input type expected by the successor. The final node produces either a ranked paper list, an evidenced paper list, or a structured graph, depending on the user query.

\noindent\textbf{Baseline fixed workflow.}
To isolate the benefit of adaptive workflow induction, we also define a fixed workflow baseline, denoted as $\Workflow{}$. This baseline uses the same toolset but executes a static pipeline for every query. A typical fixed workflow first generates search keywords from the query, retrieves papers through keyword search and citation expansion, unions and deduplicates the candidate sets, applies a generic scoring function, reranks the top candidates using an LLM, and extracts evidence for the final output. Unlike \ModelName{}, $\Workflow{}$ does not adapt its structure or parameters across search intents.

\noindent\textbf{Rollout.}
At each turn, \ModelName{} performs a workflow rollout conditioned on the current state $s_t$. The agent first induces a workflow $G_t$ by selecting operators from $\Toolset{}$ and assigning parameters according to the query, anchor paper, current candidates, and feedback history. It then executes the workflow node by node. Intermediate outputs are cached and can be reused in later turns. After execution, the agent presents the user with a ranked set of papers, supporting evidence, or a clarification question.

Formally, a rollout at turn $t$ is given by
$G_t = \Induce(q, p_0, \mathcal{P}_t, \mathcal{H}_t), (\mathcal{P}_{t+1}, y_t) = \Execute(G_t, \mathcal{P}_t),$
where $y_t$ denotes the agent's visible response, such as a ranked list, a paper comparison, or a clarification. The user then provides feedback $f_t$, and the state is updated through $\mathcal{T}$.

\noindent\textbf{Interaction-driven refinement.}
Multi-turn search is not merely repeated retrieval. It is an iterative workflow editing process: $G_{t+1} = \Refine(G_t, f_t, \mathcal{H}_{t+1}),$ where the agent updates the workflow structure, operator parameters, and intermediate candidate sets based on user feedback. This allows \ModelName{} to progressively align the search process with the user's evolving intent.

\noindent\textbf{Final output.}
After $T$ turns, \ModelName{} returns a final ranked list $\hat{\mathcal{P}}$, together with evidence explaining why each paper is relevant.

\subsection{Training}

We train \ModelName{} with a two-phase supervision pipeline that combines workflow imitation with preference optimization over corrupted workflows.
The goal is to teach the model to generate valid DAG-structured search workflows and to prefer workflows that better match the user's search intent while avoiding common structural errors.

\noindent\textbf{Training data construction.}
We start from 2,723 anchor-query training cases covering five paper-search directions.
For each case, a strong teacher model generates full search trajectories.
We extract high-quality workflow supervision examples by retaining turns where the gold paper appears in the top-5 results and satisfies the direction-specific success condition.
This yields 5,540 workflow supervision examples for supervised fine-tuning.

We then construct preference data by treating each successful workflow as the chosen response and generating rejected responses through workflow corruptions.
These corruptions cover structural validity errors and search-quality errors, including invalid references, missing inputs, incorrect operators, dropped critical nodes, shifted filters, and vague NLI axes.
After filtering easy pairs, we retain 1,733 hard chosen--rejected workflow pairs for preference optimization.

\noindent\textbf{Two-phase training.}
In the first phase, we perform supervised fine-tuning on the workflow supervision examples.
Each example is formatted as a chat-completion conversation containing the system instruction, user context, interaction history, and target assistant response.
We optimize next-token cross-entropy loss on the assistant response.

In the second phase, we continue training from the SFT checkpoint using the hard chosen--rejected workflow pairs.
Each pair shares the same input context and contrasts a successful teacher workflow with a corrupted workflow.
We optimize an IPO-style DPO objective, which provides a stable preference signal for long-context workflow generation.
Detailed training hyperparameters are provided in Appendix~\ref{app:training_details}.
\section{Experiment and Results}

\subsection{Setup}
\label{sec:setup}

We evaluate \ModelName{} on a hold-out evaluation set under both single-turn and multi-turn protocols.
In the single-turn setting, the agent searches directly from the original query.
In the multi-turn setting, it first asks clarification questions, receives simulated user feedback, and then performs retrieval using the enriched interaction context.
Full experimental details are provided in Appendix~\ref{app:experimental_setup}.

\noindent\textbf{Dataset and protocol.}
The benchmark covers five scientific search directions: \textit{predecessor}, \textit{successor}, \textit{sibling}, \textit{benchmark}, and \textit{survey}.
Each case contains an anchor paper, a user query, a search direction, and 6--15 hidden gold papers constructed from citation-graph signals, human filtering, LLM-assisted synthesis, and related-work cohorts.
We use cached Semantic Scholar metadata for each anchor paper.
For reproducible multi-turn evaluation, a fixed Qwen3.5-397B-A17B user simulator answers clarification questions across all systems.
The simulator can access hidden gold-paper metadata, but the retrieval agent never sees it.
We apply prompt constraints, deterministic string matching, and an LLM-based leak checker to prevent direct answer leakage.
Dataset, simulator, leakage-control, and inference details are provided in Appendices~\ref{app:dataset_details}--\ref{app:inference_procedure}.

\noindent\textbf{Baselines.}
We compare \ModelName{} against representative systems covering general-purpose web search, commercial deep research, fixed workflows, and tool-augmented symbolic agents.
These include GPT-5.4 with Web Search~\citep{openai2026gpt54}, OpenAI DeepResearch~\citep{openai2025deepresearch}, Qwen3.5-9B/Qwen3.5-397B~\citep{qwen2026qwen35}, GPT-5.4 with the fixed \Workflow{}, the same models with the adaptive \Toolset{}, and our trained \ModelName{}-\textit{\textsc{9B}}.
OpenAI DeepResearch is evaluated as a native one-shot deep-research system rather than a controllable workflow-refinement agent.
Detailed baseline configurations are provided in Appendix~\ref{app:baseline_details}.

\noindent\textbf{Evaluation metrics.}
Each system returns a top-50 paper list, which is compared against the hidden gold set.
We report Hit@5, Hit@10, Hit@15, Recall@50, MRR, nDCG@10, nDCG@15, average inference cost per case, and cost-efficiency.
Failed or errored executions are assigned a score of zero.
Workflow-level metrics, including TF-IDF workflow similarity and edit-consistency under add-node, modify-node, and remove-node refinements, are reported in the corresponding analysis sections.
Complete metric definitions and cost-accounting details are provided in Appendix~\ref{sec:appendix_metrics}.

\subsection{Results}
\label{sec:main_results}

\begin{table*}[t]
\centering
\scriptsize
\setlength{\tabcolsep}{2.2pt}
\renewcommand{\arraystretch}{1.12}
\begin{tabular}{
  l
  c
  l
  ccccccc
  cc
  c
}
\toprule
\textbf{Model/System}
& \multicolumn{2}{c}{\textbf{Configuration}}
& \multicolumn{7}{c}{\textbf{Retrieval Quality}}
& \multicolumn{2}{c}{\textbf{Cost-effectiveness}}
& \textbf{Cost} \\
\cmidrule(lr){2-3}
\cmidrule(lr){4-10}
\cmidrule(lr){11-12}
\cmidrule(lr){13-13}
& \textbf{Turns}
& \textbf{Tool Setup}
& \textbf{Hit@5}
& \textbf{Hit@10}
& \textbf{Hit@15}
& \textbf{R@50}
& \textbf{MRR}
& \textbf{nDCG@10}
& \textbf{nDCG@15}
& \textbf{CE$\uparrow$}
& \textbf{Rel. Eff.$\uparrow$}
& \textbf{\$/case} \\
\midrule

\multirow{6}{*}{GPT-5.4}
& 1 & Web Search
& 72.5 & 79.0 & 81.0 & 44.0 & 60.2 & 33.5 & 35.4 & \ceraw{2.0} & \releff{1.00} & 0.3703 \\
& 1 & Workflow
& 52.0 & 70.0 & 83.0 & 54.1 & 36.4 & 18.8 & 22.1 & \ceraw{2.9} & \releff{1.10} & 0.1803 \\
& 1 & Web Search + Toolset
& 79.5 & 82.0 & 84.5 & 53.5 & 62.8 & 37.1 & 39.4 & \ceraw{2.6} & \releff{1.07} & 0.3115 \\
& M & Web Search
& 79.0 & 85.0 & 89.5 & 48.2 & 65.3 & 35.3 & 37.5 & \ceraw{5.6} & \releff{1.00} & 0.1404 \\
& M & Workflow
& 56.5 & 78.0 & 84.0 & 53.6 & 38.2 & 20.5 & 24.0 & \ceraw{2.5} & \releff{0.82} & 0.2231 \\
& M & Web Search + Toolset
& 84.0 & 87.0 & 89.5 & 56.8 & 71.8 & 41.6 & 43.8 & \ceraw{5.6} & \releff{1.00} & 0.1508 \\

\midrule
o4-mini DeepResearch
& 1$^\ast$ & Web Search
& 72.0 & 81.5 & 84.5 & 46.3 & 53.0 & 29.2 & 31.7 & \ceraw{0.1} & \releff{0.47} & 6.0903 \\

\midrule
\multirow{4}{*}{Qwen3.5-9B}
& 1 & Workflow
& 48.5 & 66.5 & 77.0 & 53.7 & 31.2 & 17.4 & 21.5 & \ceraw{147.0} & \releff{2.93} & 0.0033 \\
& 1 & Toolset
& 69.0 & 82.0 & 85.5 & 38.3 & 56.5 & 31.2 & 33.4 & \ceraw{168.3} & \releff{3.03} & 0.0041 \\
& M & Workflow
& 48.0 & 67.5 & 75.0 & 52.6 & 32.7 & 17.4 & 20.4 & \ceraw{88.9} & \releff{2.00} & 0.0054 \\
& M & Toolset
& 58.0 & 71.0 & 75.5 & 34.8 & 47.5 & 26.8 & 29.0 & \ceraw{43.3} & \releff{1.67} & 0.0134 \\

\midrule
\multirow{4}{*}{Qwen3.5-397B}
& 1 & Workflow
& 70.0 & 78.0 & 82.0 & 41.1 & 53.6 & 28.4 & 29.7 & \ceraw{30.0} & \releff{1.97} & 0.0233 \\
& 1 & Toolset
& 74.0 & 87.0 & 89.5 & 43.2 & 57.8 & 34.1 & 37.0 & \ceraw{15.7} & \releff{1.67} & 0.0470 \\
& M & Workflow
& 82.0 & 85.0 & 86.5 & 42.1 & 67.9 & 35.5 & 36.0 & \ceraw{12.5} & \releff{1.22} & 0.0657 \\
& M & Toolset
& 78.5 & 88.5 & 91.5 & 48.6 & 62.1 & 38.0 & 41.6 & \ceraw{7.3} & \releff{1.07} & 0.1082 \\

\midrule
\rowcolor{paperpilotpink!18}
\ModelName{}-\textit{\textsc{9B}}
& M & Workflow + Toolset
& 77.0 & 83.5 & 89.5 & 40.0 & 59.4 & 32.5 & 35.4 & \ceraw{42.8} & \releff{1.66} & 0.0180 \\
\bottomrule
\end{tabular}

\caption{Main experiment results on the hold-out retrieval benchmark. Rows are grouped by model/system, with \textbf{Turns} indicating the interaction setting ($1$ = single-turn, $M$ = multi-turn, and $1^\ast$ = one user request handled by the native DeepResearch system).
\textbf{CE} denotes the number of successful retrieval cases per dollar, where success is defined as $\mathrm{Hit@5}=1$.
\textbf{Rel. Eff.} denotes the log-scale normalized CE score.
}
\label{tab:main_results}
\end{table*}

Table~\ref{tab:main_results} reports the main retrieval results on the hold-out evaluation set.
Overall, adaptive workflow induction improves scientific literature search over fixed workflows, and multi-turn interaction is most effective when user feedback can be translated into concrete workflow edits.
For GPT-5.4, using \ModelName{}-\textsc{Toolset} instead of the fixed \ModelName{}-\textsc{Workflow} improves single-turn Hit@5 from 52.0 to 79.5 and MRR from 36.4 to 62.8.
A similar trend appears for Qwen3.5-9B, where the toolset improves single-turn Hit@5 from 48.5 to 69.0.
These gains show that static retrieval pipelines are insufficient for diverse scientific search intents, while query-specific workflows substantially improve retrieval quality.

\begin{figure}[t]
    \centering
    \includegraphics[width=\linewidth]{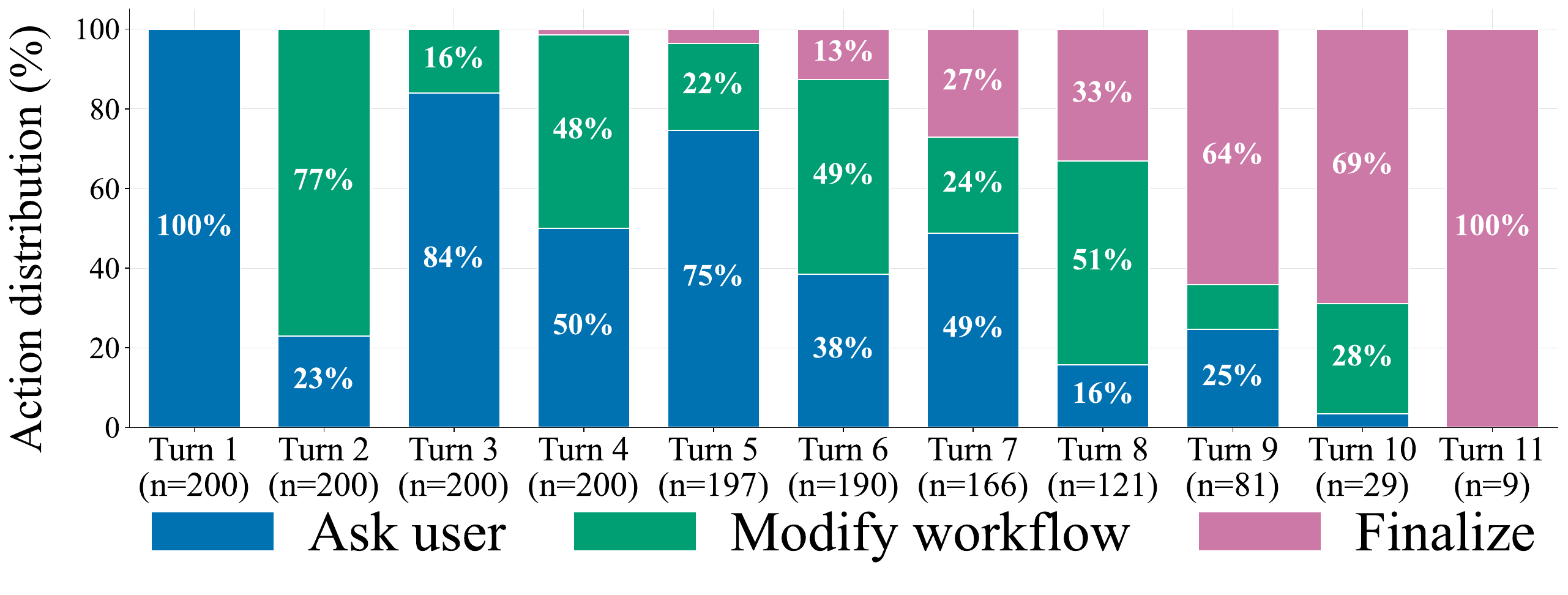}
    \caption{
    Distribution of agent actions across turns.
    }
       \vspace{-0.1in}
    \label{fig:turn_actions}
\end{figure}

Figure~\ref{fig:turn_actions} illustrates how agent behavior evolves across interaction turns.
Early turns are primarily used to gather missing user intent and search constraints, while later turns increasingly focus on workflow refinement and finalization.
This suggests that the agent separates clarification, workflow editing, and retrieval finalization into different interaction stages.

Multi-turn interaction further improves strong tool-using agents.
GPT-5.4 with Web Search improves from 72.5 to 79.0 Hit@5 and from 60.2 to 65.3 MRR under multi-turn interaction.
The best overall result is achieved by GPT-5.4 with Web Search and \ModelName{}-\textsc{Toolset} in the multi-turn setting, reaching 84.0 Hit@5, 87.0 Hit@10, 89.5 Hit@15, 56.8 Recall@50, 71.8 MRR, and 41.6 nDCG@10.
These results suggest that clarification feedback is most useful when it can be grounded into concrete search operations, such as modifying keywords, changing filters, or reranking with updated criteria.

\begin{figure}[t]
    \centering
    \includegraphics[width=\linewidth]{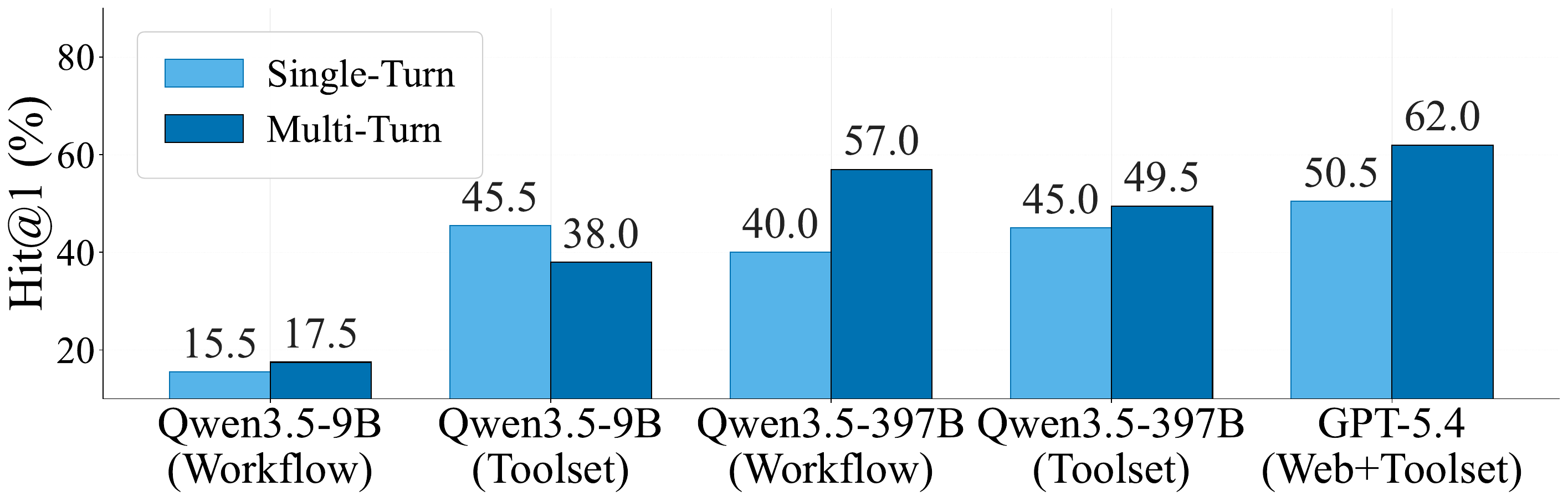}
    \caption{
    Effect of multi-turn refinement on retrieval quality across representative systems.
    }
    \vspace{-0.2in}
    \label{fig:mt_hit1}
\end{figure}

However, multi-turn interaction also introduces workflow-editing challenges for smaller untrained models.
As shown in Figure~\ref{fig:mt_hit1}, Qwen3.5-9B with \ModelName{}-\textsc{Toolset} drops under multi-turn interaction, with workflow execution errors increasing from 2.0\% to 9.5\%.
After workflow-induction training, \ModelName{}-9B reduces the workflow execution error rate to 0\%, indicating that the proposed supervision pipeline improves workflow-editing stability.

\begin{figure}[t]
    \centering
    \includegraphics[width=\linewidth]{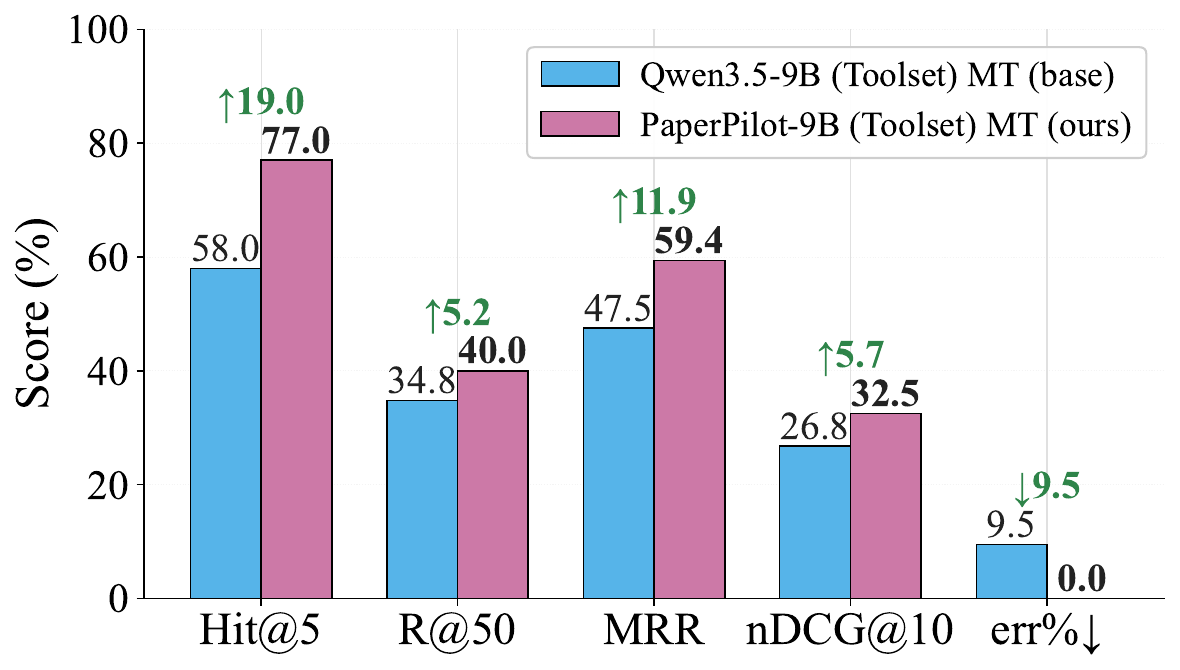}
    \caption{
     Compared with the base Qwen3.5-9B \ModelName{}-\textsc{Toolset} agent, \ModelName{}-9B improves retrieval quality across Hit@5, Recall@50, MRR, and nDCG@10, while reducing the workflow execution error rate from 9.5\% to 0\%.
    }
    \vspace{-0.2in}
    \label{fig:training_effect}
\end{figure}

Our trained \ModelName{}-9B is competitive with much larger systems.
It achieves 77.0 Hit@5, 83.5 Hit@10, 89.5 Hit@15, 40.0 Recall@50, 59.4 MRR, and 32.5 nDCG@10 in the multi-turn setting.
Compared with the base Qwen3.5-9B \ModelName{}-\textsc{Toolset} agent, it improves Hit@5 from 58.0 to 77.0, Recall@50 from 34.8 to 40.0, MRR from 47.5 to 59.4, and nDCG@10 from 26.8 to 32.5, while reducing workflow execution errors from 9.5\% to 0\%.
Figure~\ref{fig:training_effect} summarizes these gains.
Notably, \ModelName{}-9B also surpasses the single-turn GPT-5.4 Web Search baseline on Hit@5, Hit@10, Hit@15, and MRR, despite using a smaller model and lower inference cost.

Finally, \ModelName{} provides a favorable cost-performance tradeoff.
OpenAI DeepResearch costs 6.0903 dollars per case while achieving 72.0 Hit@5 and 53.0 MRR.
In contrast, GPT-5.4 with Web Search and \ModelName{}-\textsc{Toolset} achieves the best overall retrieval quality at 0.1508 dollars per case, and \ModelName{}-9B substantially improves over the base 9B toolset agent while remaining two orders of magnitude cheaper than DeepResearch.
These results highlight the practical importance of structured tool use and workflow control for scientific search.
\section{Discussion}

\begin{table*}[t]
\centering
\scriptsize
\setlength{\tabcolsep}{2.8pt}
\renewcommand{\arraystretch}{1.12}

\begin{tabular}{lccccccc|cccccccc}
\toprule
\textbf{Model}
& \multicolumn{7}{c|}{\textbf{Workflow Generation}}
& \multicolumn{8}{c}{\textbf{ Workflow Editing}} \\

\cmidrule(lr){2-8}
\cmidrule(lr){9-16}

& \multicolumn{4}{c}{\textbf{TF-IDF Cosine}}
& \multicolumn{2}{c}{\textbf{Cost-eff.}}
& \textbf{Cost}
& \multicolumn{2}{c}{\textbf{Add}}
& \multicolumn{2}{c}{\textbf{Modify}}
& \multicolumn{2}{c}{\textbf{Remove}}
& \multicolumn{2}{c}{\textbf{Full}} \\

\cmidrule(lr){2-5}
\cmidrule(lr){6-7}
\cmidrule(lr){8-8}
\cmidrule(lr){9-10}
\cmidrule(lr){11-12}
\cmidrule(lr){13-14}
\cmidrule(lr){15-16}

& Whole
& Query
& Filter
& Sig.
& CE$\uparrow$
& Rel.Eff.$\uparrow$
& \$/case
& $\Delta$cos
& $\Delta$jacc
& $\Delta$cos
& $\Delta$jacc
& $\Delta$cos
& $\Delta$jacc
& cos
& cosQ \\

\midrule

Qwen3.5-9B
& 0.0470 & 0.0190 & 0.0317 & 0.0281
& 99 & 1.94 & 0.00071
& 0.4243 & 0.2083
& 0.3586 & 0.2028
& 0.4097 & 0.4093
& 0.9382 & 0.8676 \\

Kimi K2.6
& 0.5080 & 0.2280 & 0.4000 & 0.3870
& 69 & 1.83 & 0.00965
& 0.1380 & 0.0160
& 0.2900 & 0.0970
& 0.1570 & 0.1560
& 0.5560 & 0.4160 \\

GPT-5.4
& 0.7010 & 0.3250 & 0.5400 & 0.4960
& 7 & 1.00 & 0.12565
& 0.1910 & 0.0050
& 0.4330 & 0.0940
& 0.0150 & 0.0140
& 0.8250 & 0.4720 \\

\midrule

\rowcolor{paperpilotpink!18}
\ModelName{}-\textit{\textsc{9B}}
& 0.3619 & 0.1308 & 0.2310 & 0.2238
& \textbf{665} & \textbf{3.26} & 0.00108
& \textbf{0.4303} & \textbf{0.2423}
& 0.3676 & \textbf{0.2226}
& \textbf{0.5394} & \textbf{0.5493}
& \textbf{0.9469} & \textbf{0.8777} \\

\bottomrule
\end{tabular}

\caption{
Workflow-level evaluation on generation and editing tasks.
\textbf{CE} denotes successful workflow-generation cases per dollar, and \textbf{Rel. Eff.} denotes the log-scale normalized CE score.
}
\label{tab:workflow_level_eval}

\end{table*}

\subsection{Ablation Study on Workflow Induction}
\label{sec:ablation_workflow}

To assess whether \ModelName{} improves symbolic workflow construction and refinement, we conduct an additional ablation study. Unlike the main evaluation of final ranked lists, this ablation evaluates intermediate workflow-generation behavior.

\subsubsection{Setup}

We evaluate workflow-level behavior with two complementary tasks: \textit{workflow induction} and \textit{step-wise workflow refinement}.
In workflow induction, the model receives the full interaction context and directly generates the final executable DAG, testing whether it can synthesize user intent, anchor-paper metadata, and multi-turn feedback into a functional workflow.
In step-wise workflow refinement, the model receives the current DAG state and the latest feedback context, and outputs the next DAG after applying the corresponding workflow edit.
For this workflow-level evaluation, we additionally include Kimi K2.6~\citep{moonshot2026kimi26} as a proprietary baseline for comparison.

We compare generated workflows against reference workflows using graph-level and text-level similarity metrics.
For workflow induction, we report TF-IDF cosine similarity over the whole workflow, query fields, filter predicates, and function signatures.
For refinement, we report similarity improvements after editing and final full-workflow similarity.
Full task definitions and metric details are provided in Appendix~\ref{app:workflow_metrics}.

\subsubsection{Results}

\begin{figure}[t]
    \centering
    \includegraphics[width=\linewidth]{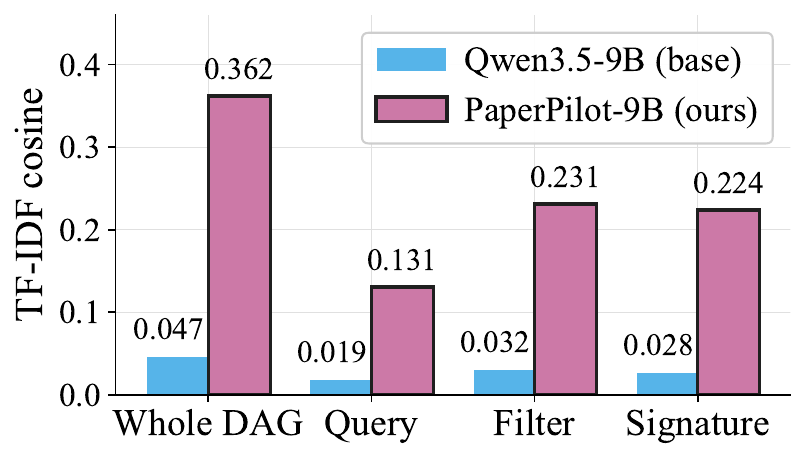}
    \vspace{-0.1in}
    \caption{
    Workflow-generation similarity between Qwen3.5-9B and \ModelName{}-9B.
    }
    \vspace{-0.1in}\label{fig:workflow_generation}
\end{figure}

\begin{figure}[t]
    \centering
    \includegraphics[width=\linewidth]{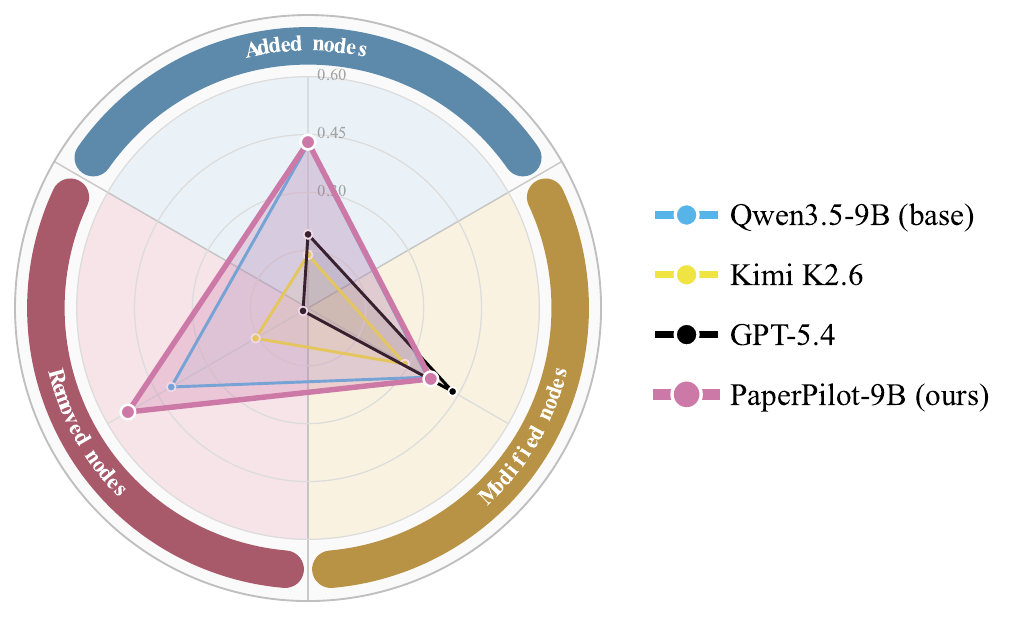}
    \caption{
    Workflow-editing consistency across different refinement operations.
    \ModelName{}-9B achieves the strongest overall local-edit stability, especially under add-node and remove-node refinements.
    }
     \vspace{-0.1in}
    \label{fig:workflow_editing}
\end{figure}

Table~\ref{tab:workflow_level_eval} reports the workflow-level evaluation results.
The left block evaluates workflow generation, while the right block evaluates workflow editing.

For workflow generation, GPT-5.4 achieves the highest absolute similarity, reaching 0.7010 whole-workflow cosine similarity and 0.4960 signature similarity, but at substantially higher cost.
In contrast, \ModelName{}-9B provides a stronger cost-effectiveness frontier, achieving a relative efficiency score of 3.26 at only \$0.00108 per case.
Compared with the base Qwen3.5-9B model, \ModelName{}-9B improves whole-workflow similarity from 0.0470 to 0.3619 and signature similarity from 0.0281 to 0.2238, indicating that workflow-induction training helps the model produce structured DAG workflows rather than shallow template-like outputs.

For workflow editing, \ModelName{}-9B achieves the strongest overall refinement performance, with the highest final workflow cosine similarity (0.9469) and query-level cosine similarity (0.8777).
It also improves over the base Qwen3.5-9B model across add-node, modify-node, and remove-node refinements, with the largest gain under remove-node editing.
Figure~\ref{fig:workflow_editing} further shows that \ModelName{}-9B maintains strong local-edit consistency across different refinement operations.

Overall, \ModelName{}-9B can both generate workflows from scratch and apply targeted structural refinements after user feedback.
These results support the central claim that workflow-induction training helps the model treat user feedback as workflow-level editing instructions rather than merely as additional query text.

\subsection{Sensitivity Analysis on Search Scale}
\label{sec:sensitivity}

We analyze whether increasing the first-stage candidate pool improves final retrieval quality.
We vary the candidate-pool size from $K=8$ to $K=20$, corresponding to $1.0\times$--$2.5\times$ the base search scale, while keeping the model, interaction protocol, and evaluation set fixed.

\begin{figure}[t]
\centering
\includegraphics[width=1.0\linewidth]{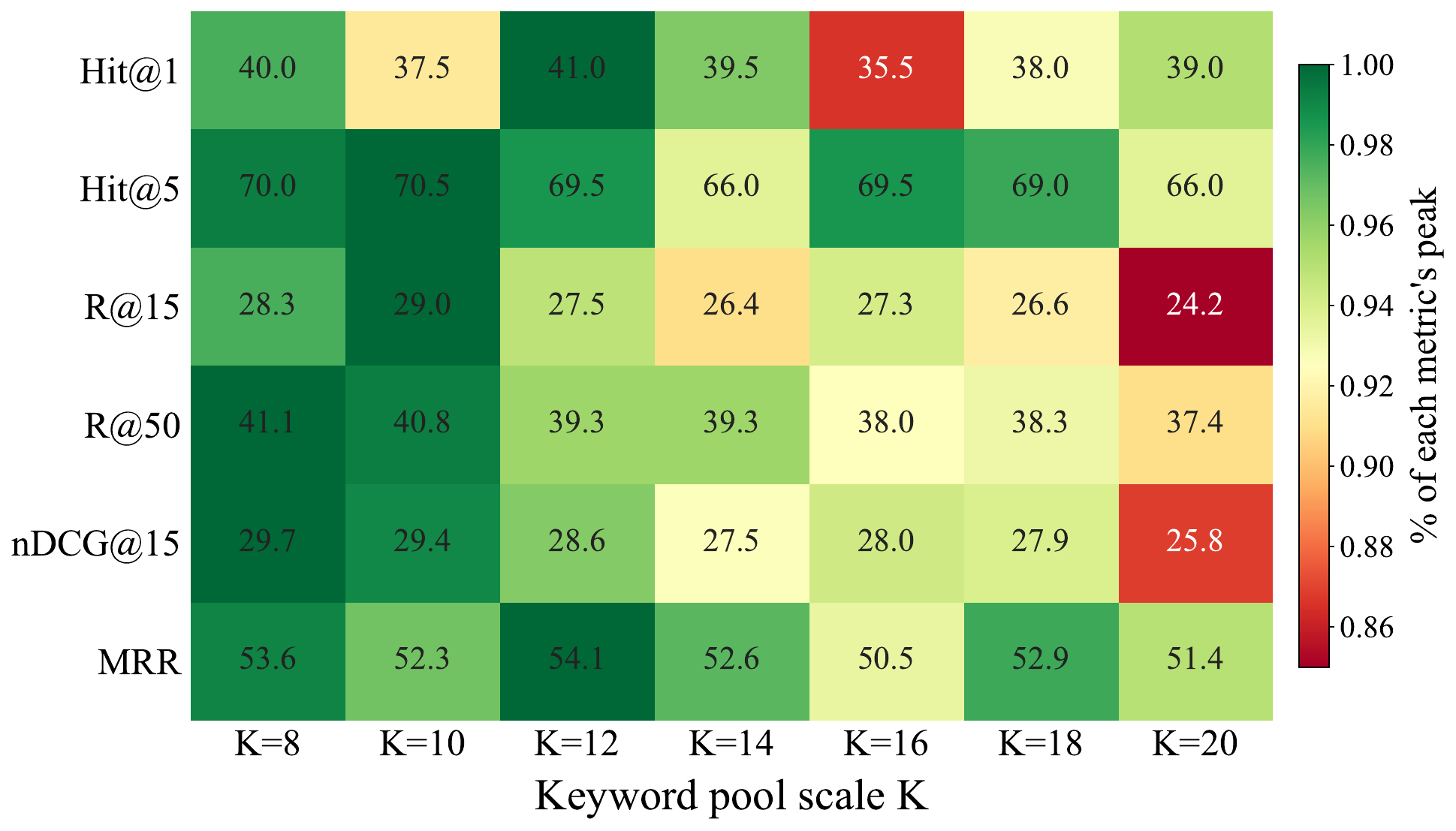}
\caption{
Sensitivity analysis over keyword pool scales.
Each cell reports the metric value at a given pool scale, with color normalized by each metric's peak.
Retrieval quality is strongest around $K=8$--$10$ and generally degrades as the candidate pool grows.
}
\vspace{-0.1in}
\label{fig:scaling_law}
\end{figure}

Figure~\ref{fig:scaling_law} and Appendix Table~\ref{tab:sensitivity_scale} show that larger candidate pools do not consistently improve retrieval quality.
Most metrics peak or remain near their peak at $K=8$--$10$, while larger pools generally reduce performance.
For example, Recall@50 drops from $0.411$ at $K=8$ to $0.374$ at $K=20$, and Hit@5 drops from $0.705$ at $K=10$ to $0.660$ at $K=20$.
This suggests that larger pools introduce more distractors, weakening downstream filtering and reranking.

\subsection{Human Study}

\begin{table}[ht]
\centering
\scriptsize
\setlength{\tabcolsep}{5pt}
\renewcommand{\arraystretch}{1.4}
\label{tab:results}
\begin{tabular}{>{\bfseries}lcccc}
\toprule
\color{black}\textbf{} &
\color{black}\textbf{SR (\%)} $\uparrow$ &
\color{black}\textbf{T1D} $\downarrow$ &
\color{black}\textbf{QSS} $\uparrow$ &
\color{black}\textbf{NTS} $\downarrow$ \\
\midrule
\text{GPT-5.4}    & 32.0          & 7.8          & 2.4          & 4.0 \\
\text{OpenAI-DeepResearch} & 8.0           & 27.4         & --          & 1.0 \\
\rowcolor{paperpilotpink!18}
\text{\ModelName{}} & \textbf{74.7} & \textbf{2.4} & \textbf{4.2} & 3.8 \\
\bottomrule
\end{tabular}
\caption{
Human-study results averaged over five sessions per participant.
SR = Success Rate (\%); T1D = Top-1 Distance (lower is better); QSS = Question Satisfaction Score (0--5); NTS = Number of Turns Until Satisfaction (lower is better).
Best values are in \textbf{bold}.
}
\label{tbl:human}
 \vspace{-0.1in}
\end{table}

We additionally conduct a human study in which we recruit participants to interact with each system across five sessions, evaluating performance along four metrics. The (1) \texttt{Success Rate} (SR, \%) measures the percentage of retrieved paper that the user found satisfactory per session. (2) The \texttt{Top-1 Distance} (T1D) captures the rank distance between the user's preferred result and the system's top-ranked item, where lower values are better. (3) \texttt{Question Satisfaction Score} (QSS, 0--5) is a user-reported rating of overall question-asked-by-agent quality. Finally, the (4) \texttt{Turns Until Satisfaction} (NTS) counts the number of conversational turns required before the user was satisfied; a low value accompanied by zero satisfaction likely reflects early abandonment rather than genuine efficiency.

As shown in Table~\ref{tbl:human}, \ModelName{} substantially outperforms both baselines across all primary metrics, achieving a success rate of 74.7\%, a mean top-1 distance of 2.4, and a question satisfaction score of 4.2. \texttt{GPT-5.4} reaches a moderate success rate of 32.0\% and a satisfaction score of 2.4, but falls well short of \ModelName{}. \texttt{OpenAI-DeepResearch} performs poorly overall, with a success rate of only 8.0\%. 
Although it records the fewest turns until satisfaction (1.0), this is because it is evaluated as a one-shot system rather than a multi-turn agent.
\section{Conclusion}

We presented \ModelName{}, a multi-turn literature search agent that formulates scientific search as workflow induction. 
Instead of relying on fixed pipelines or implicit language-only reasoning, \ModelName{} constructs executable DAG-structured workflows over typed paper-search operators and refines them through user feedback. 
With supervised workflow imitation and preference optimization over corrupted workflows, \ModelName{}-\textit{\textsc{9B}} learns to generate and edit valid search workflows. 
Our experiments show that \ModelName{}-\textit{\textsc{9B}} improves over the base 9B toolset agent under multi-turn interaction, while reducing workflow execution errors from 9.5\% to 0\%. 
It also provides a favorable cost-performance tradeoff compared with larger proprietary systems.
These results suggest that explicit, editable workflows provide a practical interface for aligning scientific literature search agents with evolving user intent.

\section*{Limitations}

\ModelName{} has several limitations. 
First, it relies on a predefined symbolic operator library, which may not cover all domain-specific search behaviors or specialized research workflows. 
Second, the workflow supervision data are generated from teacher-model trajectories and may inherit biases or blind spots from the teacher system. 
Third, our benchmark focuses primarily on computer science literature and uses controlled hidden-gold evaluation, which enables reproducible comparison but may not fully reflect open-ended real-world literature search. 
Finally, the multi-turn setting relies on an LLM-based user simulator for scalability and reproducibility. 
Although we apply leakage-control mechanisms, simulated feedback cannot fully replace human interaction. 
Future work should evaluate \ModelName{} across broader scientific domains, real users, and more adaptive operator libraries.

\section*{Use of LLMs}
 In this work, LLMs are used strictly for research support rather than as sources of substantive content. Their use falls into: (i) serving as the tested and trained model, and (ii) assisting with language refinement during paper writing. For writing support, we used GPT-5 solely to polish text (improving coherence and grammar) while all ideas, logic, results, and technical contributions originate from the authors.

\bibliography{custom}

\appendix

\section{Appendix}
\label{sec:appendix}

\subsection{Toolset}
\label{app:toolset}

Table~\ref{tab:symbolic-operator-registry} summarizes the paper-search operator library used by \ModelName{}, including sourcing, filtering, scoring, reranking, LLM-based keyword generation, and graph construction operators.

\begin{table*}[t]

\centering

\footnotesize

\setlength{\tabcolsep}{3pt}

\begin{tabular}{c l l l}

\toprule

\# & Operator & Inputs & Params / Output \\

\midrule

\multicolumn{4}{l}{\textbf{A. Sourcing}} \\

1 & keyword\_search & — & query, top\_k $\to$ PaperSet \\

2 & citation\_expand & papers & direction, max $\to$ PaperSet \\

3 & web\_resolve & — & urls $\to$ PaperSet \\

\midrule

\multicolumn{4}{l}{\textbf{B. Combine / Filter}} \\

4 & union & list[PaperSet] & $\to$ PaperSet \\

5 & dedupe & PaperSet & $\to$ PaperSet \\

6 & filter & PaperSet & predicate + params $\to$ PaperSet \\

\midrule

\multicolumn{4}{l}{\textbf{C. Scoring}} \\

7 & score & PaperSet & features, formula $\to$ Scored \\

\midrule

\multicolumn{4}{l}{\textbf{D. Cutting}} \\

8 & top\_k & PaperSet & k $\to$ subset \\

9 & above & Scored & feature $\geq$ value \\

\midrule

\multicolumn{4}{l}{\textbf{E. Rerank}} \\

10 & llm\_rerank & PaperSet & top\_k, query $\to$ PaperSet \\

11 & nli\_filter & PaperSet & categories $\to$ Classified \\

12 & fine\_read & PaperSet & top\_k $\to$ PaperSet \\

\midrule

\multicolumn{4}{l}{\textbf{F. LLM}} \\

13 & llm\_keywords & — & n, query $\to$ Keywords \\

14 & llm\_keywords\_from & PaperSet & n $\to$ Keywords \\

\midrule

\multicolumn{4}{l}{\textbf{G. Output}} \\

15 & extract\_evidence & PaperSet & $\to$ Evidenced \\

16 & pairwise\_nli & PaperSet & top\_k $\to$ Edges \\

17 & build\_graph & Papers + Edges & $\to$ Graph \\

\bottomrule

\end{tabular}

\caption{Toolset.}

\label{tab:symbolic-operator-registry}

\end{table*}

\subsection{Full Task Formulation}
\label{app:task_formulation}
We formulate multi-turn agentic paper search as an interactive decision-making problem between an agent and a human user. The task is initialized with an \emph{anchor paper} $p_0$ and a \emph{user query} $q$, where $q$ specifies the user's search intent, such as finding follow-up work, identifying strong baselines, comparing related methods, or exploring adjacent research directions. Starting from $p_0$, the agent iteratively expands, filters, and ranks a candidate paper set to satisfy the information need expressed by $q$.

At each turn $t$, the agent observes a state $s_t$ and selects an action $a_t \sim \pi(\cdot \mid s_t)$. 
The state consists of the query $q$, the anchor paper $p_0$, the current candidate set $\mathcal{P}_t$, and the interaction history
\[
\mathcal{H}_t = \{(a_1, f_1), \dots, (a_{t-1}, f_{t-1})\},
\]
where $f_i$ denotes the user feedback received after action $a_i$, including relevance judgments, preference updates, clarification responses, or refinement instructions. 
The action space is structured around a set of paper-search operators, allowing the agent to retrieve new papers, expand citation neighborhoods, combine candidate sets, filter papers by constraints, score and rerank results, or ask the user for clarification.

Each action, together with the resulting feedback, induces a state transition
\[
s_{t+1} = \mathcal{T}(s_t, a_t, f_t),
\]
which updates both the candidate set and the interaction history. After a finite horizon $T$, the agent outputs a ranked paper list $\hat{\mathcal{P}}$. The objective is to identify a target set of papers $\mathcal{P}^*$ that best satisfies the user's query while minimizing unnecessary search operations and user interactions. We therefore view the task as maximizing expected cumulative utility:
\[
\max_{\pi} \; \mathbb{E}_{\pi}
\left[
\sum_{t=1}^{T} r(s_t, a_t, f_t)
\right],
\]
where the reward captures paper relevance, alignment with user feedback, coverage of the requested search intent, and interaction efficiency.

\section{Related Work}
\label{sec:related_work}

\paragraph{Scientific literature search agents.}
A growing line of work develops LLM-based agents specialized for scientific literature retrieval and synthesis. LitLLM~\citep{agarwal2024litllm} drafts related-work sections from a user abstract by retrieving and summarizing candidate papers. PaperQA2~\citep{skarlinski2024language} performs agentic retrieval over full-text PDFs and produces citation-grounded answers. OpenScholar~\citep{asai2024openscholar} retrieves over a 45M-paper open-access corpus and synthesizes evidence-grounded scientific responses. AutoSurvey~\citep{wang2024autosurvey} and STORM~\citep{shao2024assisting} target long-form synthesis (surveys and Wikipedia-style articles) by drafting subsections from retrieved evidence. ResearchAgent~\citep{baek2025researchagent} and the AI Scientist~\citep{lu2024aiscientist} use citation-grounded literature as a substrate for idea generation and end-to-end scientific discovery. ChatCite~\citep{li2024chatcite} performs comparative literature summarization through a reflective incremental mechanism, while PaSa~\citep{he2025pasa} and SPAR~\citep{li2025spar} address comprehensive academic paper search through crawler--selector duos and multi-agent retrieval pipelines, respectively. Commercial tools such as Elicit~\citep{whitfield2023elicit} provide column-wise structured extraction over large scholarly corpora. Despite their progress, these systems typically execute a fixed pipeline or rely on free-form reasoning traces, making the retrieval process difficult for users to inspect or steer at the level of individual search operations.

\paragraph{Agentic workflow induction.}
A parallel line of work studies how LLM agents construct or follow structured workflows. ReAct~\citep{yao2023react} and Toolformer~\citep{schick2023toolformer} establish flat sequential interleavings of reasoning and tool use, while Reflexion~\citep{shinn2023reflexion} adds verbal self-reflection over the resulting trajectories. More recent frameworks induce explicit graph-structured programs of tool calls: DSPy~\citep{khattab2024dspy} compiles typed-module DAGs into self-improving pipelines, and AFlow~\citep{zhang2025aflow} discovers code-represented workflows over a typed operator library using Monte Carlo tree search. \ModelName{} draws on this line of work but differs in two respects: (i) it induces and edits a workflow conditioned on an evolving user dialogue rather than a fixed benchmark task, and (ii) its operators are paper-search primitives (citation expansion, evidence extraction, NLI filtering, etc.) rather than general code or text-transformation modules.

\paragraph{Multi-turn and interactive retrieval.}
Earlier work in conversational and clarification-driven retrieval has explored asking clarifying questions to disambiguate user intent~\citep{chi2024clarinet, li2026pearl}. General-purpose deep-research systems such as OpenAI DeepResearch~\citep{openai2025deepresearch} and Search-o1~\citep{li2025searcho1} integrate iterative search with large-scale reasoning models, but treat the retrieval procedure as a black box. In contrast, \ModelName{} translates user clarifications into concrete edits over an explicit workflow graph, providing an inspectable interface between dialogue feedback and retrieval behavior.

\paragraph{Comparison.}
Table~\ref{tab:related_work_comparison} summarizes how representative literature-search and agentic-reasoning systems differ from \ModelName{} along five key capabilities: symbolic workflow induction, workflow refinement, multi-turn dialogue, citation-graph expansion, and evidence grounding. To avoid conflating loosely related abilities (e.g., verbal self-reflection vs.\ explicit workflow editing, or citation-backed synthesis vs.\ paper-level evidence snippets), we adopt strict per-column definitions and use $\triangle$ for partial or indirect support and $-$ for capabilities that fall outside a system's design scope. Under these strict definitions, \ModelName{} is the only compared system that provides explicit first-class support for all five capabilities.

\section{Experimental Setup Details}
\label{app:experimental_setup}

\subsection{Dataset Details}
\label{app:dataset_details}

Each case contains the fields \texttt{case\_id}, \texttt{anchor\_paper\_id}, \texttt{anchor\_title}, \texttt{arxiv\_id}, \texttt{user\_query}, \texttt{search\_direction}, \texttt{golden\_paper\_ids}, \texttt{golden\_titles}, and \texttt{n\_gold}.
The benchmark covers five search directions: \textit{predecessor}, \textit{successor}, \textit{sibling}, \textit{benchmark}, and \textit{survey}.
These directions correspond to common scientific search needs: finding prior work, follow-up work, methodologically similar papers, benchmark or dataset papers, and survey papers.

Each case contains 6--15 gold papers.
The gold labels are constructed from the anchor paper's citation graph, human filtering, LLM-assisted synthesis, and intra-related-work cohorts.
Before running each case, we pre-fetch anchor metadata from Semantic Scholar, including the abstract, publication year, citation count, and TLDR when available.
The metadata is cached locally and used as part of the agent input.

\subsection{User Simulation and Leakage Control}
\label{app:user_simulation}

Multi-turn evaluation requires user feedback, but collecting human feedback for every model and every case is expensive and difficult to reproduce.
We therefore use an LLM-based user simulator in the multi-turn setting.
The simulator is conditioned on the anchor metadata, original user query, search direction, hidden gold paper metadata, conversation history, and the current clarification question with candidate options.
It then role-plays the user by either selecting one of the provided options or producing a short free-form answer.

We use Qwen3.5-397B-A17B as the simulator for all multi-turn experiments.
The simulator is fixed across all systems, ensuring that different agents receive feedback from the same user model.
Importantly, the simulator has access to the hidden gold papers, but the search agent does not.
This allows the simulator to provide preference-aware feedback while preserving the hidden-answer evaluation protocol.

To prevent answer leakage, we enforce a strict leakage-control pipeline.
First, the simulator system prompt explicitly instructs the model not to mention gold paper titles, authors, or other identifying information.
Second, we apply a deterministic string-matching detector, \texttt{detect\_leak()}, which checks whether the simulator response contains substrings from gold paper titles or author names.
Third, we use a Qwen3.5-9B leak-checker model to judge whether the simulator response reveals gold information in a less direct form.
If a response is flagged as leaking information, the simulator is asked to regenerate.
If repeated regeneration still leaks information, the response is replaced with a generic preference-level answer.
This process ensures that the simulated user can guide the agent through natural feedback without directly revealing the answer.

\subsection{Inference Procedure}
\label{app:inference_procedure}

In the single-turn setting, each case is evaluated independently.
We first retrieve the anchor paper metadata from Semantic Scholar.
The agent then runs one retrieval pass and returns the top-50 papers.
For fixed-workflow systems, this pass follows a deterministic pipeline consisting of first-stage sourcing, scoring, and reranking.
For toolset-based systems, the agent generates a symbolic workflow DAG and executes it over the paper-search backend.
The final top-50 list is compared against the hidden gold set to compute retrieval metrics.

In the multi-turn setting, we add a clarification loop before retrieval.
The gold metadata is provided only to the user simulator and never to the retrieval agent.
At each clarification turn, the clarifier LLM asks one question with 3--5 answer options.
The simulator responds from the user's perspective, and the response is passed through the leakage-control pipeline described above.
The resulting question-answer pair is appended to the interaction history.
The loop continues until the clarifier outputs \texttt{finalize} or reaches the maximum number of clarification turns, which is set to 4--5 depending on the experiment.
We then concatenate the original query and accumulated Q/A history into an enriched query.
The retrieval agent runs once with this enriched query and outputs the top-50 papers.

\subsection{Baseline Details}
\label{app:baseline_details}

We compare against eight major system configurations across single-turn and multi-turn settings.
The evaluated systems include general-purpose web-search agents, commercial deep-research systems, fixed workflow baselines, and tool-augmented symbolic agents.

\paragraph{Web search.}
We evaluate GPT-5.4 with a standard web-search tool.
This baseline represents a strong general-purpose search agent that can issue natural-language queries and synthesize results, but does not use the \Toolset{}.

\paragraph{DeepResearch.}
We include OpenAI DeepResearch as a strong commercial scientific search baseline.
Because its API only supports single-turn requests in our evaluation setting and does not expose a controllable workflow-refinement interface, we evaluate it as a one-shot baseline.
Given the same anchor-paper metadata and user query, it is asked to return a ranked list of relevant papers.
Thus, DeepResearch serves as a commercial one-shot search reference rather than a directly comparable multi-turn workflow-refinement agent.

\paragraph{\Workflow{}.}
We evaluate Qwen3.5-9B, Qwen3.5-397B, and GPT-5.4 with a fixed \ModelName{} workflow.
This baseline uses the same paper-search backend, but follows a deterministic pipeline for every query.
The pipeline consists of first-stage sourcing, scoring, and reranking.
Because the workflow structure is fixed, this baseline isolates the benefit of tool use from the benefit of adaptive workflow induction.

\paragraph{\Toolset{}.}
We evaluate Qwen3.5-9B, Qwen3.5-397B, and GPT-5.4 with the full \ModelName{} symbolic toolset.
In this setting, the model can generate a DAG-structured workflow using available paper-search operators, configure their parameters, and execute the workflow.
Unlike \ModelName{}-9B, these models are not trained with our workflow induction and preference optimization pipeline.

\paragraph{\ModelName{}-9B.}
Finally, we evaluate our trained \ModelName{}-9B model.
It is trained to generate and refine DAG-structured workflows conditioned on the anchor paper, user query, interaction history, and simulated user feedback.
This setting tests whether specialized workflow-induction training improves multi-turn scientific search beyond prompting a large model with tools.

\subsection{Evaluation Metrics}
\label{sec:appendix_metrics}

All metrics are averaged over cases in the hold-out evaluation set.
We also report finer-grained analysis by search direction, grouping cases into \textit{predecessor}, \textit{successor}, \textit{sibling}, \textit{benchmark}, and \textit{survey}.
We use strict first-pass scoring: failed or errored cases are assigned a score of zero rather than being skipped.

\paragraph{Hit@$K$.}
Hit@$K$ measures whether at least one gold paper appears in the top-$K$ returned papers:
\[
\mathrm{Hit@}K = \mathbb{I}\left[\hat{P}_{1:K} \cap P^* \neq \emptyset\right],
\]
where $\hat{P}_{1:K}$ is the top-$K$ retrieved list and $P^*$ is the hidden gold set.
We report Hit@5, Hit@10, and Hit@15 in the main results.
These metrics measure whether the system can surface at least one relevant paper within a small review set, reflecting early-stage retrieval usefulness for literature search.

\paragraph{Recall@$K$.}
Recall@$K$ measures how many gold papers are recovered in the top-$K$ results:
\[
\mathrm{Recall@}K =
\frac{|\hat{P}_{1:K} \cap P^*|}{\min(K, |P^*|)}.
\]
We report Recall@50 in the main table and use additional cutoffs for analysis when needed.
This metric captures coverage when each query has multiple relevant gold papers.

\paragraph{MRR.}
Mean reciprocal rank measures how early the first gold paper appears:
\[
\mathrm{MRR} =
\frac{1}{N}
\sum_{i=1}^{N}
\frac{1}{r_i},
\]
where $r_i$ is the rank of the first gold paper for case $i$.
If no gold paper appears in the returned list, the reciprocal rank is zero.

\paragraph{nDCG@$K$.}
We report normalized discounted cumulative gain to measure rank-sensitive retrieval quality:
\[
\mathrm{DCG@}K =
\sum_{j=1}^{K}
\frac{\mathrm{rel}_j}{\log_2(j+1)},
\]
\[
\mathrm{nDCG@}K =
\frac{\mathrm{DCG@}K}{\mathrm{IDCG@}K}.
\]
Here, $\mathrm{rel}_j=1$ if the paper at rank $j$ is in the gold set and $0$ otherwise.
We report nDCG@10 and nDCG@15 in the main table.


\paragraph{Workflow-generation metrics.}
For workflow induction, we compute TF-IDF cosine similarity over the whole workflow, query-related fields, filter predicates, and function signatures.
For workflow refinement, we report cosine and Jaccard similarity changes under add-node, modify-node, and remove-node edits, together with final full-workflow similarity.
These metrics evaluate whether a model can generate and locally refine executable DAG workflows rather than only producing the final retrieved paper list.

\paragraph{Cost and cost-effectiveness.}
We report the average dollar cost per case and cost-effectiveness metrics.
Cost is computed from agent-side token usage only.
Tokens from the user simulator and leak checker are excluded because they are part of the evaluation infrastructure rather than the deployed user-facing system.
We compute dollar cost using public list prices for each model and embedding endpoint.

Cost-effectiveness (CE) measures the number of successful cases achieved per dollar.
For the main retrieval benchmark, a case is counted as successful if at least one gold paper appears in the top-5 results, i.e., $\mathrm{Hit@5}=1$.
Thus, CE is computed as the number of successful retrieval cases divided by the total inference cost.
For the workflow-generation task, a case is counted as successful according to the workflow-generation success criterion, and CE is computed analogously as successful workflow-generation cases per dollar.
We also report relative efficiency (Rel. Eff.), a log-scale normalized CE score, to compare systems with substantially different cost ranges.
Higher CE and Rel. Eff. indicate better cost-performance tradeoffs.

\subsection{Training Details}
\label{app:training_details}

For supervised fine-tuning, we train for 3 epochs with learning rate $2\times10^{-4}$ and sequence length 14,336.
We use LoRA adaptation over both attention and MLP projections.

For preference optimization, we continue from the SFT checkpoint and optimize an IPO-style DPO objective with $\beta=0.2$.
The preference stage is trained for 3 epochs with learning rate $3\times10^{-5}$, sequence length 16,384, and gradient clipping with maximum norm 1.0.

\subsection{Workflow-Level Evaluation Metrics}
\label{app:workflow_metrics}

We evaluate workflow-generation behavior by comparing each generated DAG with a reference workflow. 
We use both graph-level and text-level metrics to capture structural correctness and parameter-level alignment.

\paragraph{Graph-level metrics.}
We compute Jaccard similarity over node identifiers and operator types. 
Node-identifier Jaccard measures whether the generated workflow preserves the reference workflow structure, while operator-type Jaccard measures whether it selects similar paper-search operations regardless of exact node names. 
We also report edit-distance statistics, including the number of added, removed, and modified nodes required to transform the generated workflow into the reference workflow. 
Lower edit distance indicates more stable workflow construction.

\paragraph{Text-level metrics.}
We serialize each workflow into normalized text fields and compute TF-IDF cosine similarity against the reference workflow. 
We report cosine similarity over the whole workflow, query-related fields, filter predicates, and function signatures. 
Whole-workflow similarity measures overall alignment; query-field similarity captures whether the model preserves the intended search formulation; filter similarity evaluates constraint-level alignment; and signature similarity measures whether the generated operators expose compatible input-output behavior.

\paragraph{Refinement metrics.}
For step-wise workflow refinement, we evaluate whether each edit moves the current workflow closer to the reference next-state workflow. 
We report similarity changes after editing and final full-workflow similarity.
Specifically, given the current workflow $G_t$, the edited workflow $G_{t+1}$, and the reference next-state workflow $G^{*}_{t+1}$, we compute:
\[
\begin{aligned}
\Delta \mathrm{cos} &=
\mathrm{cos}(G_{t+1}, G^{*}_{t+1})
-
\mathrm{cos}(G_t, G^{*}_{t+1}), \\
\Delta \mathrm{jacc} &=
\mathrm{Jacc}(G_{t+1}, G^{*}_{t+1})
-
\mathrm{Jacc}(G_t, G^{*}_{t+1}).
\end{aligned}
\]
Positive values indicate that the refinement improves alignment with the reference workflow. 
We also report the final full-workflow cosine similarity after all refinement steps.

\subsection{Human Study Details}
We conduct the human study to evaluate how well each system supports interactive paper search in realistic multi-turn settings. We recruit six volunteers from different majors (Biology, Social Science, Computer Science) and different academic levels (Undergraduate, Master Student, PhD student). Participants interact with each system over five sessions, and after each session we collect both outcome-based and user-reported metrics. We evaluate four dimensions: \texttt{Success Rate} (SR, \%), defined as the percentage of retrieved papers that the user considers satisfactory; \texttt{Top-1 Distance} (T1D), which measures the rank distance between the user-preferred paper and the system's top-ranked result, with lower values indicating better ranking alignment; \texttt{Question Satisfaction Score} (QSS, 0--5), a subjective rating of the quality and usefulness of the agent's clarification questions; and \texttt{Turns Until Satisfaction} (NTS), the number of conversational turns required before the user reaches a satisfactory result. We note that low NTS should be interpreted together with SR, since early termination without satisfaction may indicate abandonment rather than efficient interaction.

\begin{figure*}
    \centering
    \includegraphics[width=0.95\textwidth]{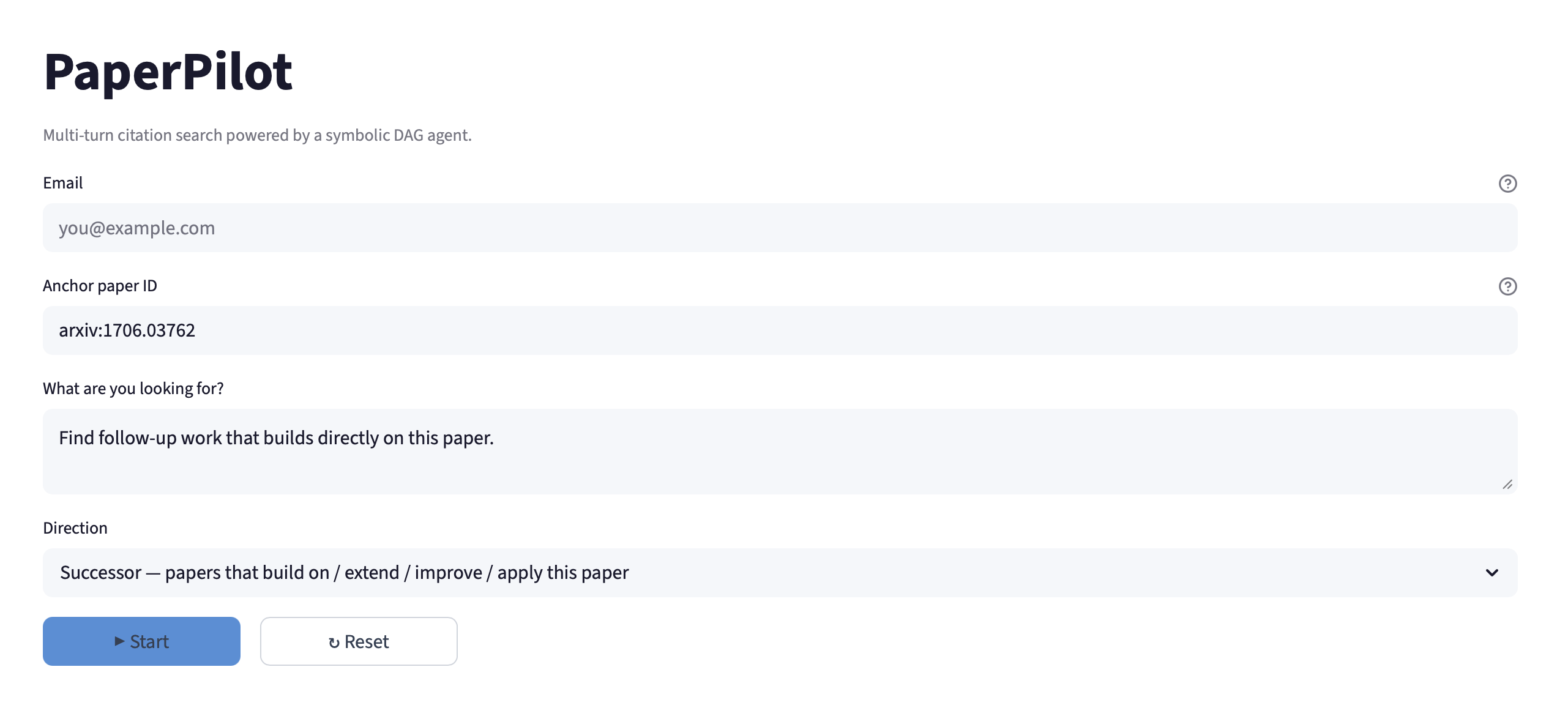}
    \caption{Human Study interface}
    \label{fig:placeholder}
\end{figure*}

\begin{figure*}
    \centering
    \includegraphics[width=0.95\textwidth]{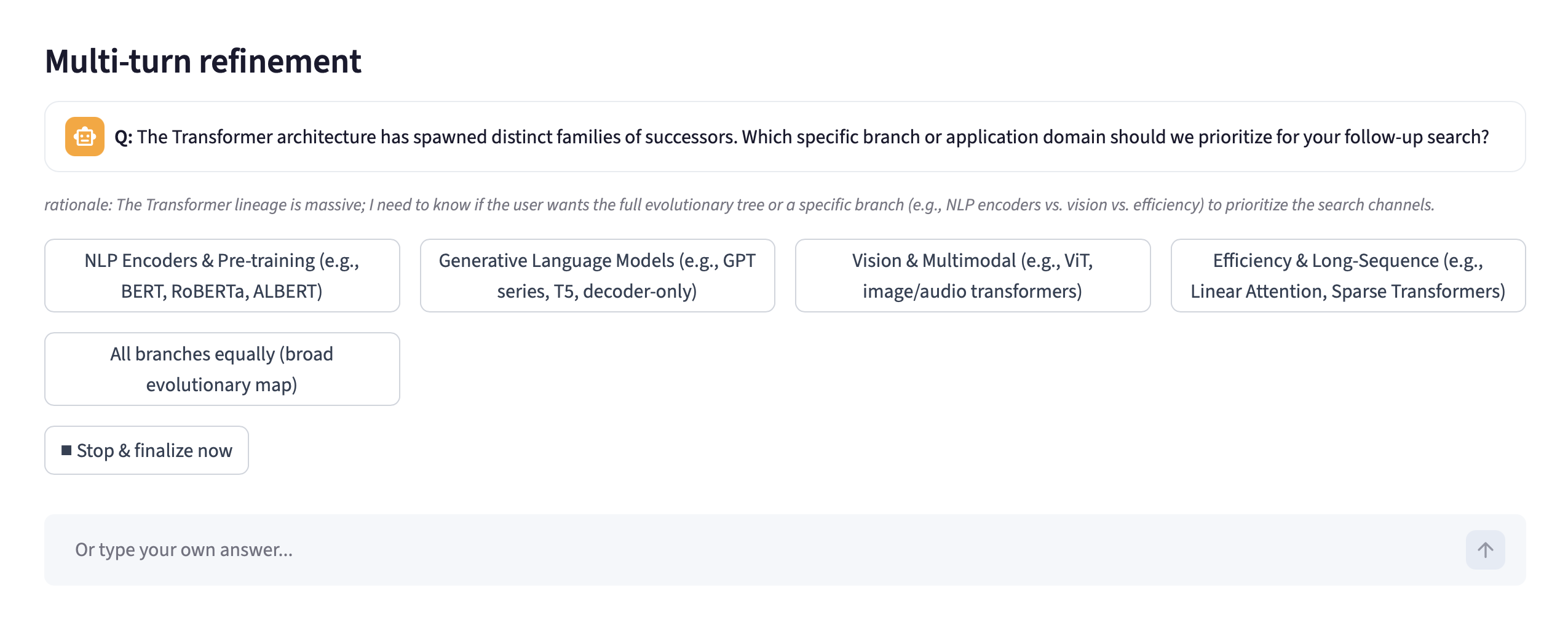}
    \caption{Human Study interface}
    \label{fig:placeholder}
\end{figure*}

\begin{figure*}
    \centering
    \includegraphics[width=0.95\textwidth]{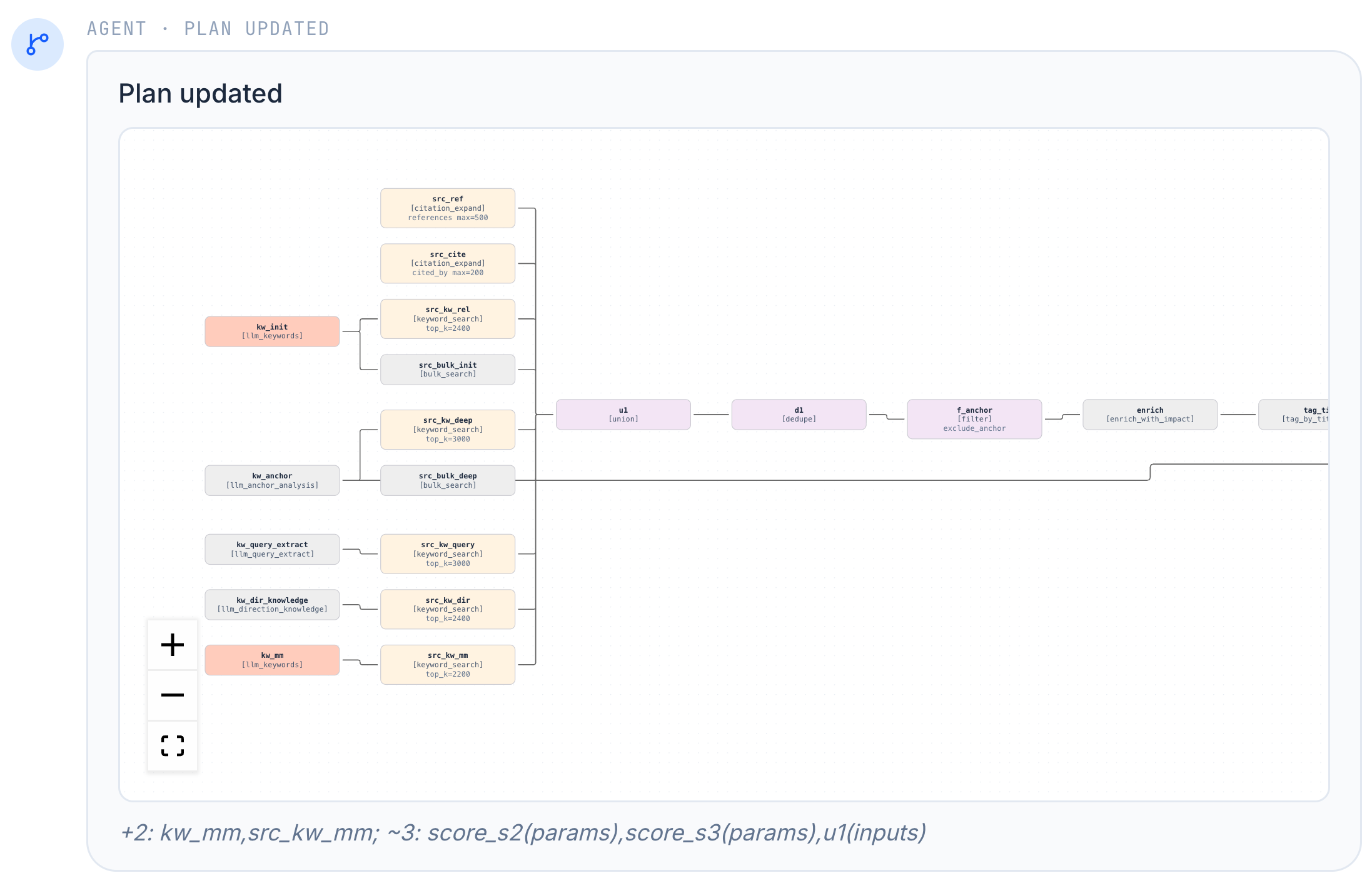}
    \caption{Workflow refinement example}
    \label{fig:placeholder}
\end{figure*}

\begin{figure*}
    \centering
    \includegraphics[width=0.95\textwidth]{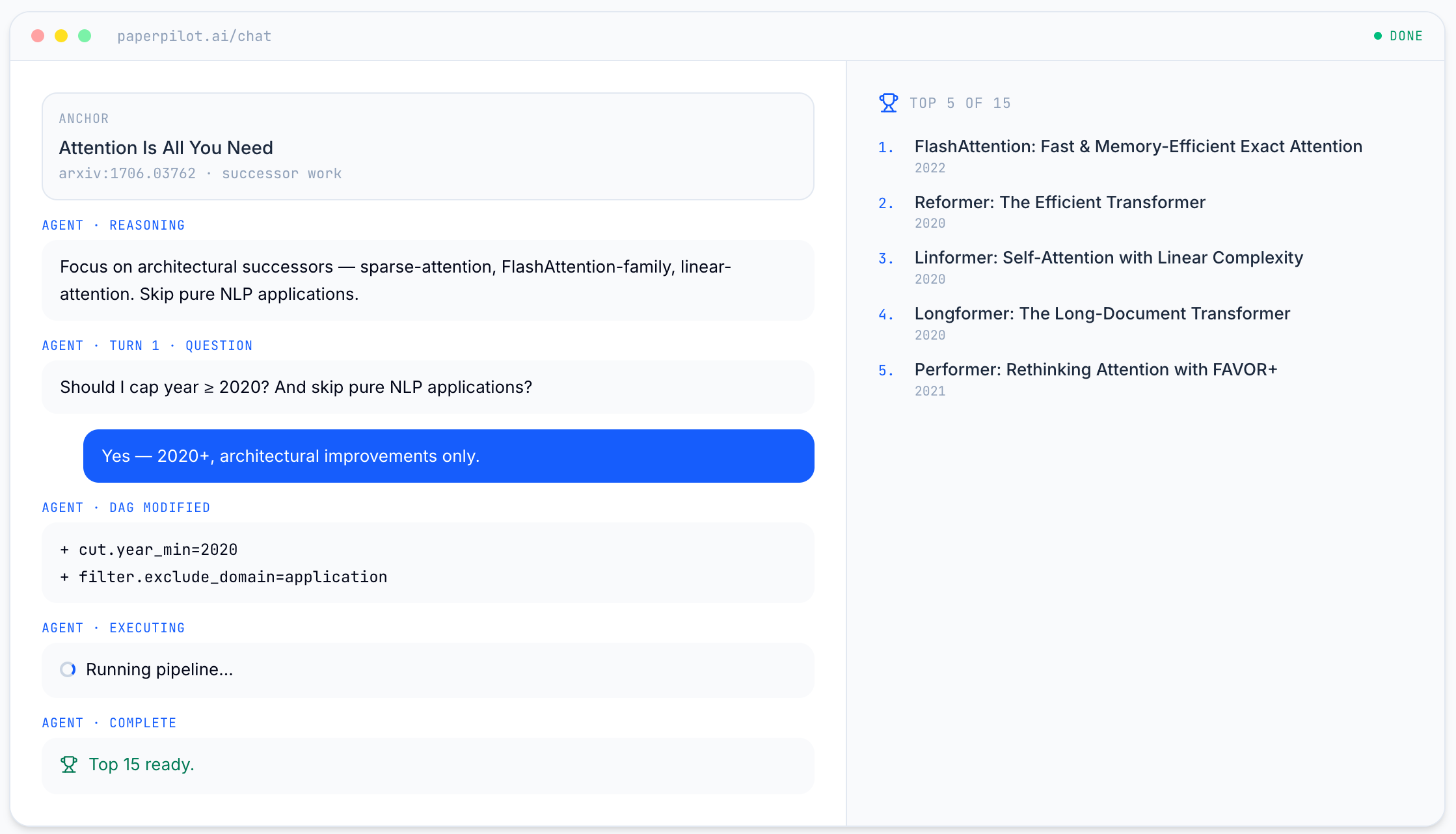}
    \caption{An example PaperPilot search session}
    \label{fig:placeholder}
\end{figure*}

\section{Sensitivity Analysis Results}

\begin{table}[H]
\centering
\scriptsize
\setlength{\tabcolsep}{3.0pt}
\renewcommand{\arraystretch}{1.08}
\begin{tabular}{lccccccc}
\toprule
\textbf{Variant} 
& \textbf{H@1} 
& \textbf{H@5} 
& \textbf{R@15}
& \textbf{R@50}
& \textbf{nD@15} 
& \textbf{MRR} 
& \textbf{Pool} \\
\midrule
K8  (1.0$\times$)  & .400 & .700 & .283 & .411 & .297 & .536 & 4,472  \\
K10 (1.25$\times$) & .375 & .705 & .290 & .408 & .294 & .523 & 5,549  \\
K12 (1.5$\times$)  & .410 & .695 & .275 & .393 & .286 & .541 & 6,612  \\
K14 (1.75$\times$) & .395 & .660 & .264 & .393 & .275 & .526 & 7,658  \\
K16 (2.0$\times$)  & .355 & .695 & .273 & .380 & .280 & .505 & 8,550  \\
K18 (2.25$\times$) & .380 & .690 & .266 & .383 & .279 & .529 & 9,540  \\
K20 (2.5$\times$)  & .390 & .660 & .242 & .374 & .258 & .514 & 10,523 \\
\bottomrule
\end{tabular}
\caption{
Sensitivity analysis on search scale. Increasing the first-stage candidate pool does not monotonically improve final retrieval quality; larger pools often introduce more distractors and reduce recall-oriented metrics.
}
\label{tab:sensitivity_scale}
\end{table}

\end{document}